\documentclass{article}

\usepackage{arxiv}
\usepackage{amsmath}
\usepackage[utf8]{inputenc} 
\usepackage[T1]{fontenc}    
\usepackage{hyperref}       
\usepackage{url}            
\usepackage{booktabs}       
\usepackage{amsfonts}       
\usepackage{nicefrac}       
\usepackage{microtype}      
\usepackage{cleveref}       
\usepackage{lipsum}         
\usepackage{graphicx}
\usepackage{natbib}
\usepackage{doi}
\usepackage{amssymb}
\usepackage{float}
\floatstyle{plaintop}
\restylefloat{table}

\usepackage{breqn}
\usepackage{soul, color}
\usepackage{graphicx}

\title{Deep-MDS Framework for Recovering the 3D Shape of 2D Landmarks from a Single Image}

\author{Shima Kamyab \\ 
	Department of Computer Engineering \\
	Shiraz University \\
	Fars, Iran \\
	\texttt{sh.kamyab@cse.shirazu.ac.ir} \\
	\And
	Zohreh Azimifar \\
	Department of Computer Engineering\\
	Shiraz University\\
	Fars, Iran \\
	\texttt{azimifar@cse.shirazu.ac.ir} \\
}

\hypersetup{
pdftitle={Deep-MDS Framework for Recovering the 3D Shape of 2D Landmarks from a Single Image},
pdfsubject={Deep Learning, 3D reconstruction},
pdfauthor={Shima Kamyab, Zohreh Azimifar},
pdfkeywords={Single view 3D human face shape recovery, deep learning human face shape recovery, using multidimensional scaling for 2D shape recovery from a singe image},
}

\begin{document}
\maketitle

\begin{abstract}
	In this paper, a low parameter deep learning framework utilizing the Non-metric Multi-Dimensional scaling (NMDS) method, is proposed to recover the 3D shape of 2D landmarks on a human face, in a single input image. Hence, NMDS approach is used for the first time  to establish a mapping from a 2D landmark space to the corresponding 3D shape space. A deep neural network learns the pairwise dissimilarity among 2D landmarks, used by NMDS approach, whose objective is to learn the pairwise 3D Euclidean distance of the corresponding 2D landmarks on the input image. This scheme results in a symmetric dissimilarity matrix, with the rank larger than 2, leading the NMDS approach toward appropriately recovering the 3D shape of corresponding 2D landmarks. In the case of posed images and complex image formation processes like perspective projection which causes occlusion in the input image, we consider an autoencoder component in the proposed framework, as an occlusion removal part, which turns different input views of the human face into a profile view. The results of a performance evaluation using different synthetic and real-world human face datasets, including Besel Face Model (BFM), CelebA, CoMA - FLAME, and CASIA-3D, indicates the comparable performance of the proposed framework, despite its small number of training parameters, with the related state-of-the-art and powerful 3D reconstruction methods from the literature, in terms of efficiency and accuracy.
\end{abstract}


\section{Introduction}
\label{sec:Introduction}

 In general, human face 3D reconstruction plays a fundamental role in many computer vision applications such as face recognition, gaming, etc. Utilizing the 3D face shape in a framework helps increase the framework accuracy and makes it invariant to the pose and/or occlusion changes in the input~\cite{sharma2020voxel, sadeghzadeh2020pose, komal2020technical}. On the other hand, this problem is a computationally demanding process which is considered a limitation in some environments like embedded systems.

One effective solution to reduce the complexity of a 3D reconstruction framework, is using geometric features, called landmarks instead of the whole input image.  Landmarks are well-known and efficient features which are applicable in many computer vision tasks, such as camera calibration~\cite{bartl2020planecalib, rehder2017online}, mensuration~\cite{naqvi2020measurement}, image registration~\cite{zhang2020headlocnet, bhavana2020medical}, scene reconstruction~\cite{nouduri2020deep, ngo2021micp}, object recognition~\cite{choi2020facial, bocchi2020object}, and motion analysis~\cite{bandini2020new, malti2021exact}. The main advantage of landmark-based approaches is that they are less sensitive to lightning conditions or other radiometric variations~\cite{rohr2001landmark}. As stated in~\cite{tian2018landmark}, 2D geometric information, like landmarks and contours, sometimes are more effective than photometric stereo-based for 3D reconstruction in the wild, while photometric information, including lighting, camera properties, and reflectance properties, are themselves \textbf{less robust} features especially for unconstrained images. Nowadays, several accurate landmarks detection algorithms and methods in the literature use a different standard. ~\cite{king2009dlib, zhu2012face, zou2020towards, FLAME:SiggraphAsia2017}. 

Using simple features like landmarks, with efficient use of memory and computation, makes the solution suitable for the high-latency and energy-efficient embedded devices~\cite{chen2021virtual,wisth2021unified}, where the objective is to push the computing closer to where the sensors gather data. Therefore, the main requirement of these systems is to keep the frameworks as light as possible.

Instead of the whole image, utilizing landmark features opens up new horizons toward the point-based statistical methods like Multi-Dimensional Scaling (MDS) variants~\cite{kruskal1978multidimensional}. Formally, MDS approach refers to a set of statistical procedures used for exploratory data analysis and dimensionality reduction. It takes the estimates of similarity among a group of items, as input, or various ``indirect'' measurements (e.g., perceptual confusions), and the outcome is a ``map'' that conveys, spatially, the relationships among items, wherein similar items are located proximal to one another, and dissimilar items are located proportionately further apart~\cite{hout2013multidimensional}. A detailed description of MDS approach is given in Sec.~\ref{sec:methodology}, and, for more information, the reader is referred to~\cite{ghojogh2020multidimensional, hout2013multidimensional}. In general, MDS solution includes decomposing a dissimilarity matrix, computing from the input points, to find a mapping into a new space, preserving the input points' configuration as much as possible~\cite{ghojogh2020multidimensional}.

In this paper, the idea is to use the MDS approach for mapping a number of standard 2D landmarks on a human face, in a single input image, into the corresponding 3D shape space, by recovering the landmarks' 3D shape.  The main challenge here is that, to the best of our knowledge, the MDS approach is usually used to reduce the dimensions, not increase them. In fact, the reason why MDS reduces the dimensions is that the rank of its used dissimilarity, usually selected as the Euclidean distances among the input points, having $D$ D dimensions, is  one minus the dimensions of the input points, and therefore, MDS could map the data to at most $D-1$ dimensions.

Instead of using the Euclidean distance among the input 2D landmarks, in the MDS approach, consider using a dissimilarity matrix which gives an  \textit{estimate} of \underline{3D Euclidean distance} among the landmarks on the corresponding 3D shape. If there is a way to learn such dissimilarity, MDS approach can appropriately recover the 3D shape of the input 2D landmarks. 

We propose to learn such dissimilarity, using a deep learning framework with a pair of 2D points on a single image of a human face, as the input, and an estimate of the 3D Euclidean distance between their corresponding 3D locations, as the output. The proposed deep learning dissimilarity has then the small number of training parameters.    

From another viewpoint, the recovery of the 3D shape of some 2D landmarks in a single image, is an ill-posed problem which needs to impose some constraints, on the solution space, to obtain only \textit{feasible} solutions, for the problem at hand~\cite{engl2014inverse}. In the case of the 3D reconstruction from a single image, a usual way is to use a \textit{3D model} to constraint the resulting 3D shape to be a feasible human face, defined by the model~\cite{bfm09}. This type of constraint comes with the drawback of obtaining  \textit{biased} solutions toward the mean shape of the used 3D model~\cite{aldrian2012inverse}. In our proposed framework, instead, the needed constraints on the solution space are provided by the deep learning components and the MDS approach, which don't bias the found solutions toward a specific region of the solution space. 

In the case of different types of image formation processes, we use an autoencoder to turn any input projection type into some profile view so that the deep learning dissimilarity could be trained without any confusion. This is because we empirically observed that the self-occlusion, caused by posedness or complex projection types, like perspective, etc., mislead the performance of the deep neural network for learning the dissimilarity in our framework. In the case of the human faces, a simple autoencoder could appropriately turn the input view into an appropriate profile view.

Therefore, our contributions in this paper include:
\begin{itemize}
    \item using the MDS approach to \textit{increase} the dimensions, for the first time.
    \item proposing a deep learning symmetric dissimilarity measure to be used in the MDS approach for estimating the 3D shape of 2D landmarks in a single image.
    \item proposing a low parameter, unbiased deep learning framework for 3D recovery of the landmark locations in a single image, independent of the type of 2D projection.
\end{itemize}
 The rest of this paper is organized as follows:
Section~\ref{sec:rel_works} includes the review of the related 3D reconstruction methods to ours and their advantages and drawbacks. In Section~\ref{sec:methodology}, we demonstrate our proposed method and the concepts needed for understanding our method. We then report the experimental evaluations in Section~\ref{sec:Experiments}, followed by the Conclusions and future works, discussed in Section~\ref{sec:conclusion}.


\section{Related Works}
\label{sec:rel_works}

A large body of research in the field of 3D reconstruction has utilized spatial geometric features, like landmarks, as a cue and has reported more precise and accurate solutions. This section reviews the state-of-the-art and the most recent single view human face 3D reconstruction methods that utilize landmark depth estimation in their process.

 The approaches appeared in~\cite{zhao2017simple, chinaev2018mobileface, aldrian2012inverse, moniz2018unsupervised} are the state-of-the-art 3D reconstruction methods which use landmark depth estimation. In~\cite{zhao2017simple}, assuming the images are the result of the \textit{orthographic} projection, only the depth (i.e., the third dimension) of input 2D landmarks is estimated using a Deep Neural Network (DNN). In the case of other projection types like perspective projection, the work in~\cite{zhao2017simple} has then a poor performance. In~\cite{chinaev2018mobileface}, a low parameter deep learning framework for model-based 3D reconstruction from a single image is proposed. In this framework, the coefficients of a 3D Morphable Model (3DMM) are learned from a single input image, using a CNN. The solution is usually biased toward the 3DMM mean shape in the model-based frameworks. In~\cite{aldrian2012inverse}, a closed-form solution is proposed for estimation of the coefficients of a 3DMM from a single input image, using different assumptions about the image formation process. This method is also a model-based framework with the orthographic projection assumption. In~\cite{moniz2018unsupervised}, an unsupervised DNN is proposed to replace the face in a given image by a face in another image. To do this, a number of landmarks in the image are considered for correspondence between two images. Then, their depth, i.e., the third dimension, assuming orthographic projection, are estimated using a DNN along with a matrix for transforming the resulted 3D landmarks and aligning them with the target face pose. In this work (~\cite{moniz2018unsupervised}), the landmarks' depth is obtained along with the transformation matrix, in the output of a DNN, as an intermediate solution, and may not be a feasible landmark set by itself.

Among the recent deep learning frameworks for 3D human face reconstruction we can name~\cite{li2021detail}, which proposes a model-based 3D reconstruction of human face shape including two steps: coarse shape estimation and adding more details to the resulted face. In the coarse shape estimation phase, a deep neural network is used to estimate the 3DMM parameters from some landmarks on the input image. The estimated shape is then fine-tuned in the second phase using high frequency features obtained by a GAN. A high number of parameters and model-based schemes may be the drawbacks of these frameworks. In~\cite{wu2019cascaded}, another model-based method with a model fitting algorithm for 3D facial expression reconstruction from a single image is proposed. A cascaded regression framework is designed to estimate the parameters for the 3DMM from a single image. In this framework, 2D landmarks are detected and used to initialize the 3D shape and the mapping matrices. At each iteration, residues between the current 3DMM parameters and the ground-truth are estimated and used to update the 3D shapes. In~\cite{hu2021self}, a self-supervised 3D reconstruction framework is proposed which considers the feature representation of some landmarks in both 2D and 3D images and establishes consistency among the landmarks in all images. Using landmarks in this framework can further improve the reconstructed quality in local regions by the self-supervised classification for the landmarks. In~\cite{cai2021landmark}, a 3D reconstruction framework is proposed for caricature images. It learns a parametric model on the 3D caricature faces and trains a DNN for regressing the 3D face shape based on the learned nonlinear parametric model. The proposed DNN results in a number of detected landmarks on each input caricature face image. The proposed RingNet~\cite{RingNet:CVPR:2019} is a model-based single image 3D human face reconstruction framework without the need for 2D-3D supervision. The RingNet output is based on the FLAME model~\cite{FLAME:SiggraphAsia2017}.  RingNet leverages multiple images of a person and automatically detects 2D face features. It uses a novel loss that encourages the face shapes to be similar when the identity is the same and different for different people. In our experiments, we used this method as a recent unsupervised 3D landmark shape recovery method.

In~\cite{tian2018landmark}, a human face 3D reconstruction framework is proposed whose input includes only landmarks. This framework reconstructs a 3D face shape of the frontal pose and natural expression from multiple unconstrained images of the subject via cascade regression in 2D/3D shape space.  Using the landmarks in this framework, the 3D face shape is progressively updated, using a set of cascade regressors, which are learned offline, using a training set of unconstrained face images and their corresponding 3D shape. In~\cite{gong2015two}, it is stated that in a dense 3D reconstruction framework, knowing the depth of some landmarks on the input face is necessary to prevent the degradation of prediction accuracy. Therefore, a two-stage method is proposed which first estimates landmarks' depth and then uses them within a deformation algorithm to reconstruct a precise 3D shape. In this work, the landmarks' depth estimation needs to have two input images which are not available in many cases.

Despite their good performance in single view human face reconstruction applications, the methods reviewed above may suffer from a large number of parameters or be biased toward a mean shape. Our proposed method for 3D shape recovery of 2D landmarks could be incorporated as the landmark depth estimation component in these methods to increase their efficiency. Our proposed deep learning framework is aimed to overcome the above problems in a low parameter framework, incorporating a novel use of the analytic MDS method.

\section{Methodology}
\label{sec:methodology}

In this section, we first explain the basics of the MDS approach  based on which we designed our proposed framework. Then, we will frame the proposed method.

\subsection{Multi-dimensional Scaling (MDS) Approach}
Multidimensional Scaling (MDS) is a well-known manifold learning dimensionality reduction, and feature extraction method with several types like classical MDS, kernel classical MDS, metric MDS, and non-metric MDS. Sammon mapping and Isomap are considered as two special cases of MDS approach~\cite{ghojogh2020multidimensional}.  The MDS objective is to preserve the similarity~\cite{torgerson1965multidimensional} or dissimilarity/distances~\cite{beals1968foundations} of the points in a low-dimensional embedding space. In fact, MDS aims at preserving the local configuration of the data to capture its global structure~\cite{saul2003think}.

Formally, consider high dimensional input dataset \mbox{$\{\mathbf{x_i}\}_{i=1}^n$}, to be mapped to the embedded data \mbox{$\{\mathbf{y_i}\}_{i=1}^n$} having lower dimensions, with $n$ as the number of data points,  $x_i \in \mathbb{R}^d$, and $y_i \in \mathbb{R}^p$, and $p \leq d$. We denote $ X =[\mathbf{x_1}, ..., \mathbf{x_n}] \in \mathbb{R}^{d \times n}$ and $Y =[\mathbf{y_1}, ..., \mathbf{y_n}] \in \mathbb{R}^{p \times n}$. The goal of classical MDS is to preserve the similarity/dissimilarity of the data points while projecting from the input space $D^X$ to  the embedding space $D^Y$~\cite{torgerson1965multidimensional} as:
\begin{equation}
    minimize_{\{\mathbf{y_i}\}_{i=1}^n} c_1 = \sum_{i=1}^n \sum_{j=1}^n (D^X(i,j) - D^Y(i,j))^2
\end{equation}
If we consider dot product as the proximity measure, the solution of MDS will be \cite{ghojogh2019feature}:

\begin{equation}\label{eq:mds_solution}
    Y=\Delta^{\frac{1}{2}}V^T
\end{equation}
where we decompose $X^TX$ using eigenvalue decomposition to have
\begin{equation}\label{eq:svd}
    X^TX = V \Delta V^T
\end{equation}
Note here that the eigenvectors of $X^TX$  should be sorted from leading (largest eigenvalue) to trailing (smallest eigenvalue). Equivalently, instead of eigenvalue decomposition of $X^TX$, one
can decompose $X$ using Singular Value Decomposition (SVD) and take its \textit{leading} singular vectors. The matrix $\Delta$ is then obtained by squaring the singular values.

In summary, the classical MDS uses~\eqref{eq:svd}  to eigenvalue decompose $X^TX$, followed by using~\eqref{eq:mds_solution} to obtain $Y \in \mathbb{R}^{n \times n}$. Truncating this $Y$ to have $Y\in \mathbb{R}^{p \times n}$, with the first $p$ rows, gives a $p$-dimensional embedding of the $n$ points. This is actually equivalent to performing Principal Component Analysis (PCA) on the $X^TX$.

The classical MDS is a linear method with a closed-form solution, while the metric and non-metric MDS methods are nonlinear without closed form solutions, and should be solved iteratively. In non-metric MDS, rather than using a distance metric, $d_y(x_i, x_j )$, between two points $x_i, x_j$, in the embedding space, we use $f(d_y(x_i, x_j ))$ where $f(\cdot)$ is a non-parametric monotonic function. In other words, in non-metric MDS, only the order (not the degree) of dissimilarities is important ~\cite{agarwal2007generalized}:
\begin{equation}
    d_y(y_i,y_j) \leq d_y(y_k,y_l) 	\Leftrightarrow f(d_y(y_i,y_j)) \leq f(d_y(y_k,y_l) )
\end{equation}

The optimization in non-metric MDS is \cite{agarwal2007generalized}:
\begin{dmath}
minimize_{\{\mathbf{y_i}\}_{i=1}^n} c_2 = \left( \frac{\sum_{i=1}^n \sum_{j=1, j<i}^n (d_x(x_i,x_j)-f(d_y(y_i,y_j)))^2}{\sum_{i=1}^n \sum_{j=1, j<i}^n d_x^2(x_i,x_j)} \right)^{\frac{1}{2}}
\end{dmath}
where, some normalized difference between the distance in the input space and the embedding space is minimized. Since, the distances are symmetric the minimization only accounts for $j<i$.

\subsection{The idea}

To better understand the proposed idea of this paper, consider when we compute the Euclidean distance between the \underline{3D landmarks} on a 3D human face shape and use this dissimilarity to compute the solution of the classical MDS approach as in \eqref{eq:mds_solution}, and set $p=d=3$. In this way, the resulting embedding $Y$ will be the same as the 3D input points. 

Now assume that we could learn the \underline{Euclidean 3D distance} of the landmarks on a face from their projected \underline{2D locations}. In this way, using 2D landmarks on a 2D image, we could obtain their corresponding 3D Euclidean distance and use this proximity to recover the 3D shape of the landmarks, using the MDS approach. In this case, by definition, we do not use the classical MDS because the proximity is not the dot product of the input landmarks but an estimate of their 3D Euclidean distance. Therefore, we have a non-metric MDS approach using an estimate of the 3D Euclidean distance among the 2D input landmarks. In the following section, we will present our proposed deep learning dissimilarity network to be used in the MDS approach.

\subsection{Deep Learning Dissimilarity}
Designing a deep learning dissimilarity in our proposed framework is a simple regression of 3D Euclidean distances from 2D ones which are possible using a five-layer feed-forward neural network with less than $3000$ trainable parameters. We empirically set five feed-forward layers with the same number of neurons and a short connection between two layers to improve the achieved accuracy of the DNN for the proposed dissimilarity. 

The input to the dissimilarity is a set of different proximities between a pair of 2D landmarks $x_i, x_j$ on a human face, and the output is an estimate of their 3D Euclidean distance. Empirically, we observed that the few number of the input dimensions, i. e., two scalars for each landmark, restricts the power of this DNN so that it will be confused when the input landmarks are from a posed image or from a complex projection. Therefore, there should be an additional component to transform any input views of the human face into a profile view to be used in the dissimilarity learning process.

to preserve the symmetricity of the learned dissimilarity, we set the input to this network as
\begin{dmath}
    |x_i^1-x_j^1|,... |x_i^d-x_j^d|\\
    x_i^1x_j^1, ..., x_i^dx_j^d\\
    e^{-|x_i^1-x_j^1|},..., e^{-|x_i^d-x_j^d|}
\end{dmath}
where $d$ is the number of dimensions for each input point, and $x_i^d$ denotes the $d^{th}$ dimension for landmark  $\underline{x}_j$.

using this scheme, for the input to the dissimilarity network, we feed different forms of proximities of the input landmark pair, to obtain the corresponding 3D Euclidean distance.
\subsection{Proposed Framework}
\begin{figure}
    \centering
    \includegraphics[scale=0.6]{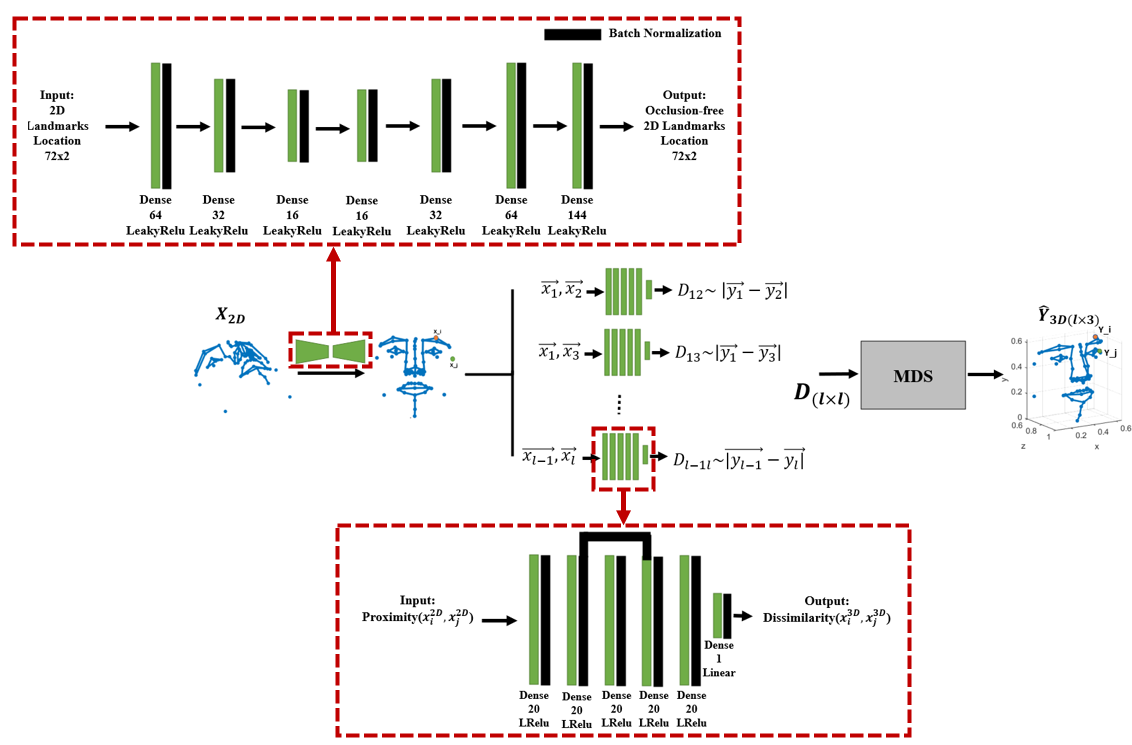}
    \caption{Diagram of proposed framework for 3D shape reconstruction from 2D landmarks in a single image.}
    \label{fig:proposed_diagram}
\end{figure}

Figure~\ref{fig:proposed_diagram} shows the diagram of our proposed framework , where, there are two separately trained deep learning components, including an autoencoder for transforming all views into a profile view, and a dissimilarity between each pair of the input 2D landmarks. The dissimilarity should be computed for all pairs of the input landmarks to compute the dissimilarity matrix,  used in the NMDS component.

For training of the autoencoder component, we used the Mean Squared Error (MSE) between the location of the 2D points in the input view and their corresponding location in the profile view as the loss function $J_{autoencoder}(\cdot)$:
\begin{equation}
    J_{autoencoder} (L_{in}^{2D}; \theta) = \sum_{i=1}^N|L_{in}^{2D, i} - L_{profile}^{2D,i}|^2
\end{equation}
where $L_{in}^{2D, i}$ denotes the $i^{th}$ landmark location in the input view, and $L_{profile}^{2D,i}$ is the location of the $i^{th}$ landmark point in the profile view, and $\theta$ stands for the DNN parameters;

In the case of the deep learning dissimilarity component, where the output should be an estimation of the 3D euclidean distance between the  corresponding 3D landmark points for an input 2D landmark pair, the loss function $J_{Dissimilarity}(\cdot)$ is as~\eqref{eq:j_dis} :
\begin{equation}\label{eq:j_dis}
    J_{Dissimilarity}(L_{in}^{2D, i}, L_{in}^{2D, j}; \gamma)=|L_{in}^{3D,i}-L_{in}^{3D,j}|^2
\end{equation}
where $\gamma$ denotes the deep learning dissimilarity parameters. 

The following section includes the experimental results for evaluation of our proposed method compared to the state-of-the-art and recent landmarks depth estimation methods, from the literature.

\section{Experiments}
\label{sec:Experiments}
In this section, we evaluate the performance of our proposed method by comparing to the state-of-the-art and recent related landmark shape recovery methods, from the literature. We perform our experiments using synthetic and real-world human face datasets. In the case of numerical comparisons, we also validate the results by performing wilcoxon signed rank statistical test, which is a well-known statistical analysis method for deep learning approaches \cite{woolson2007wilcoxon}.

We performed our experiments in two parts, based on the BFM~\cite{bfm09} and FLAME~\cite{FLAME:SiggraphAsia2017} landmark topologies. We used the datasets and methods consistent with the corresponding landmark topology in each part.
\subsection{Evaluation Criteria}
Our evaluations are based on MSE error among the 3D recovered landmarks by existing methods and its ground truth in the test set as in \eqref{eq:MSE}.
\begin{equation}\label{eq:MSE}
    MSE = \sum_{i=1}^N |L_{pred}^{3D, i} - L_{GT}^{3D,i}|
\end{equation}
where, $L_{pred}^{3D, i}, L_{GT}^{3D,i}$ denote the 3D location of $i^{th}$ predicted landmarks and its ground truth, respectively, and N is the number of landmarks.

We also used \textbf{Depth Correlation (DepthCorr)} criteria \cite{moniz2018unsupervised} which measures the correlation between the predicted landmarks and their GT. To obtain the DepthCorr value, for a prediction method, we first compute the correlation among the predicted and GT landmarks and then compute the trace of the resulting correlation matrix as DepthCorr value.

To validate the obtained results by the compared methods, we further evaluate them by performing statistical tests, to ensure that the results are significantly different. To do this, we used wilcoxon signed rank test\cite{woolson2007wilcoxon} which is a suitable statistical analysis tool for deep learning frameworks. We performed random samplings of size 500 from the test set of size at least 5000 of the used dataset, for 10 times, and evaluated the existing methods for each of the resulting 10 sampling set, using the MSE and DepthCorr criteria. In this way, we will have 10 criterion (MSE/DepthCorr) values for a specific method on the test set, and use these 10 values as the input to the wilcoxon test for a specific method.
\subsection{Experimental Setup}
We trained our method and the work in \cite{zhao2017simple} on a NVIDIA GeForce GTX 1080 GPU, with KERAS deep learning library using ADAM optimizer. In the case of analytical methods like~\cite{aldrian2012inverse}, we implement the methods, using MATLAB 2018b on a 4GHz CPU with 32 GB RAM, utilizing the available BFM model from the web.

In the case of RingNet~\cite{RingNet:CVPR:2019}, we used the public available pre-trained framework for this work which is based on the tensorflow library, implemented in Linux.

In the case of Mediapipe~\cite{lugaresi2019mediapipe} method, we used the python library available for this method.

All codes and data of our paper is publicly available for research purposes at https://github.com/s2kamyab/DeepMDS.

\subsection{Datasets}\label{sec:data}
\begin{itemize}
    \item \textbf{Besel Face Model (BFM)\cite{blanz1999morphable}} BFM is a 3D morphable model (3DMM) proposed in Computer Vision Group at the University of Basel, as a generative model for synthesizing the human faces. The geometry of the BFM consists of $53,490$ 3D vertices connected by $160,470$ triangles. Faces of different identities can be composed as linear combinations of 199 principal components.
    
    We used the BFM model with natural expression to generate 20000 random human faces with natural expression, each projected with a random pose of angle from $\{-45, 0, 45\}$ in degrees, and random projection type including orthographic and perspective, as the training set and 5000 random samples as the test set, with projection type same as the training set, for evaluation of the existing methods. In the case of perspective projection we used random values for azimuth in range $[0,45]$, elevation in range $[0,30]$ and field of view (FoV) in range $[0,5]$ for computing the view matrix in perspective projection. 
    
    Our focus in this paper is only on $72$ landmarks from BFM generated human heads and corresponding 2D projections. Therefore we extracted $72$ standard landmarks from each human head to be used in our method. However, in the case of compared model-based methods, we also provide the input 2D images.
    
    
    \item \textbf{CoMA-Registered with FLAME model \cite{COMA:ECCV18, FLAME:SiggraphAsia2017}}
    As another landmark configuration other than the BFM 3D model and its standards for the location and number of the landmarks, we also performed evaluations, using the FLAME 3D model~\cite{FLAME:SiggraphAsia2017} with $51$ standard landmarks on the 3D human faces. The COMA 3D human face dataset~\cite{COMA:ECCV18} is a 3D human face shape dataset based on the FLAME topology which contains motion sequences of $12$ extreme, asymmetric facial expressions from $12$ different subjects. In total, the dataset contains $20,466$ 3D meshes, each with about $120,000$ vertices. There are also the raw scanner data (i.e. raw scans and camera images), and temporal registrations (i.e. in FLAME topology) in the dataset. We extracted $51$ FLAME standard landmarks from each 3D face in the COMA dataset, using barycentric coordinates which are available for landmarks extraction in the FLAME model~\cite{FLAME:SiggraphAsia2017}, and projected these heads using different projection types mentioned in the BFM model description to obtain a dataset. We used the CoMA dataset only for the training  in our experiments.

    \item \textbf{CASIA3D \cite{olivetti2019deep}}
     Contains $4,624$ scans of $123$ subjects. Each subject was captured with different expressions and poses. We only use scans with expressions, which amount to $3,937$ scans over $123$ subjects. We use this dataset as the test set for  performance evaluations. To detect the 2D landmarks on CASIA3D images, we used DLib python library~\cite{king2009dlib} to detect $68$ landmarks on each face, in the input 2D image. 3D face shapes in the CASIA3D dataset are not based on the FLAME topology. Therefore, after obtaining the landmarks on the input images, using DLib, we fitted a flame 3D model to the obtained landmarks, using the method in ~\cite{FLAME:SiggraphAsia2017}. We then register the two 3D models, i. e., the CASIA3D GT model and the fitted 3D FLAME model, using the iterative closest point algorithm (ICP). After registration of the two 3D models, for each standard landmark on the fitted 3D FLAME model, the point with the smallest distance on the CASIA GT 3D face shape is found and is used as the GT landmark on the CAISIA3D GT face, consistent with the FLAME topology. By removing the first $17$ landmark points from  the above $68$ standard landmarks, the $51$ standard landmarks in our experiments are obtained.
    
    \item \textbf{CelebA \cite{liu2015faceattributes}}
    Celeb Faces Attributes (CelebA) dataset is a large-scale face attributes dataset with more than $200K$ celebrity images, each with $40$ attribute annotations. The images in this dataset cover large pose variations and background clutter. CelebA has large diversities, large quantities, and rich annotations, including $10,177$ number of identities, $202,599$ number of face images, and $5$ landmark locations, $40$ binary attributes annotations per image.  We used CelebA dataset in both parts of our experiments for visual comparisons. We used the landmarks from this dataset provided by~\cite{he2017fully, he2017robust} in our experiments based on the FLAME topology. In the case of BFM topology, we also used BFM models fitted to the CelebA dataset, using the method in~\cite{bas2016fitting} for numerical evaluations.
\end{itemize}

\subsection{Compared Methods}
We selected the compared methods from the state-of-the-art and most recent research with a landmark depth estimation step in their process including:
\begin{itemize}
\item \textbf{RingNet~\cite{RingNet:CVPR:2019}} as a recent single view model-based 3D human face shape reconstruction method based on FLAME model topology which we described in Section\ref{sec:rel_works}. RingNet output is a full 3D shape based on FLAME model topology. We only consider the 51 standard FLAME landmarks on its output in our experiments.

\item \textbf{Mediapipe~\cite{lugaresi2019mediapipe} Library for human face landmark detection} which is a python open source library with the option of 3D landmark detection on the human faces in a single input image, as a perception application. We used Mediapipe in our experiments as a recent powerful method for 3D shape recovery of the landmarks in a single image. The Mediapipe library detects $468$ 3D landmarks on the input human face. To make the Mediapipe landmarks consistent with the FLAME topology, we found the index of $51$ standard FALME landmarks on the Mediapipe output, manually. 

\item \textbf{Zhao's \cite{zhao2017simple}}, in which a deep learning framework is proposed to recover the third dimension of the landmarks in an input image, assuming the orthographic projection as the image formation process. This method is independent of a the 3D model topology, therefore we use it in both parts of our experiments.

\item \textbf{MobileFace~\cite{chen2018mobilefacenets}} includes a model-based deep learning framework for reconstruction of a full 3D human face shape, based on the BFM dataset. This framework is proposed for being used in the mobile devices, using the mobilenet DNN. There exist a new loss function defined in the shape space instead of the coefficient space and an orthographic projection term \cite{chen2018mobilefacenets}, in this method. Using BFM in this work allows to find the $72$ standard landmarks on their predictions. We choose this method as one of the compared methods in our experiments because of its claim about the low number of parameters and using BFM in their work which makes this framework similar to ours. In the experiments, we only account for the MobileFace predictions for our desired landmarks on the face and not the full 3D face shape, predicted by the MobileFace.
\item \textbf{Aldrian's~\cite{aldrian2012inverse}} is an analytic and state-of-the-art method including recovering the 3D shape of 2D landmarks on the human face in a single input image, in its framework. This work uses BFM topology in its framework with an iterative optimization approach for updating 3DMM coefficients and camera parameters, consistent with the input image, interchangeably. There is also a texture estimation process in this method. We only use 3D shape estimation part of this framework as compared method in our experiments.

\end{itemize}
\subsection{Visual and Numerical Results and Statistical Analyses}
As mentioned before, we used two landmark topologies, including BFM and FLAME, in our experiments. Therefore, we report the results for each topology, separately.
\subsubsection{BFM Landmark Topology}
In this part of our experiments, we compare our proposed framework with the existing methods, based on the BFM landmark topology. The compared methods then include: MobileFace, Aldrian's, and Zhao's.

\begin{figure}
    \centering
    \includegraphics[scale=0.5]{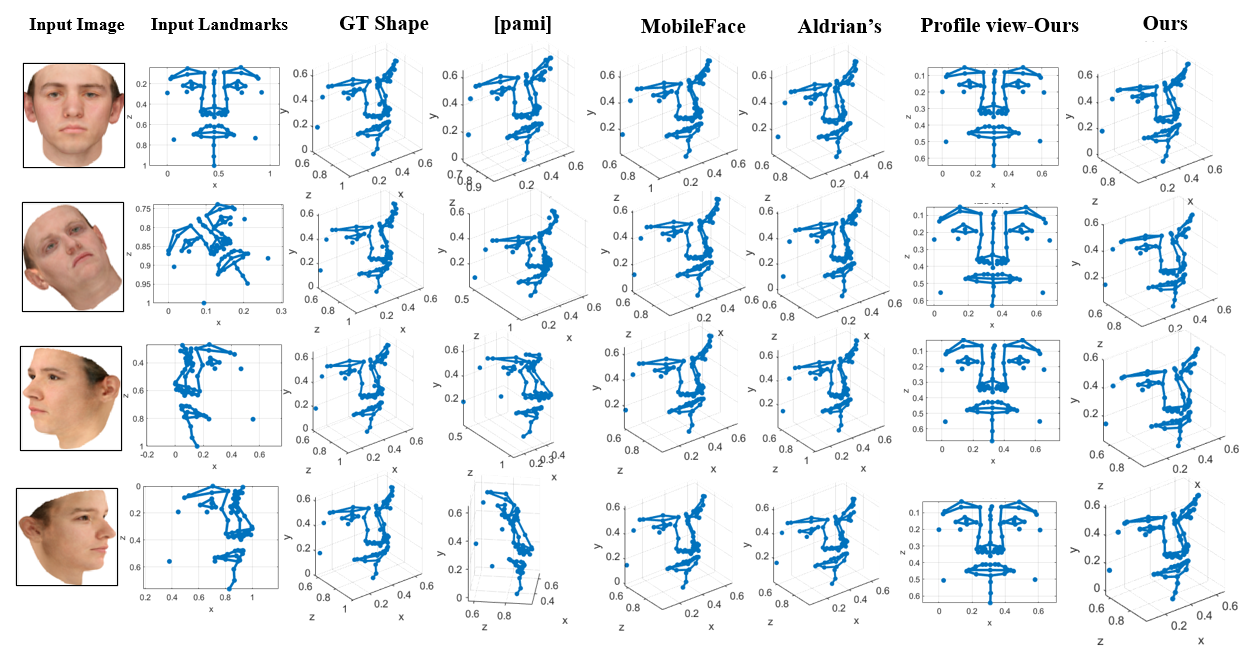}
    \caption{Visual results for pose invariant 3D shape recovery power, obtained by existing methods, using BFM topology, in our experiments.}
    \label{fig:cmp_bfm}
\end{figure}

We first compare the performance of the existing methods in the pose and projection invariant face shape recovery. To do this, we use the BFM dataset with natural expression and different 2D projections as described in section \ref{sec:data}. Figure~\ref{fig:cmp_bfm} includes the visual results obtained by different methods in this part in which we aligned the predicted shapes with the GT shape, using the Procrustes analysis~\cite{gower1975generalized} in which a moving shape is aligned to a target point cloud using rotation, translation and scale operations.

From Figure~\ref{fig:cmp_bfm} ($7^{th}, 8^{th}$ columns), it is observable that our proposed framework could suitably recover the profile view  and, accordingly, the 3D shape of the landmarks in the images with different projection types. 

\begin{table}
    \centering
    
    \begin{tabular}{|c|c|c|c|c|}
    \hline
         BFM (Synthetic)&  Zhao's & Aldrian's & MobileFace & Ours \\
         \hline
          avgDepthCorr. ($\%$)  $\uparrow$ & $71.42 \pm 0.009  $& $71.89 \pm 0.015$ & $71.97 \pm 0.002$ & $71.95 \pm 0.002$\\
        \hline
        avgMSE( $\times 10^{-4}$) $\downarrow$ & $22 \pm 0.25$&$2.5 \pm 0.18$ &$0.6 \pm 0.004$ & $1.4 \pm 0.005$ \\
        \hline

    \end{tabular}
    \caption{Numerical results obtained by existing methods in our experiments for BFM dataset.}
    \label{tab:cmp_bfm}
\end{table}

Table~\ref{tab:cmp_bfm} includes the numerical comparisons in this part of the experiments.

To provide clearer comparison, Table~\ref{tab:cmp_param} includes the number of trainable parameters and inference time complexity, for each method in our experiments. In the case of the methods which use BFM or FLAME 3D model, , due to limited access to the data structure and algorithm, we used the term $O($BFM/FLAME Model$)$, expressing the complexity of the 3D models when recovering the 3D shape from input parameters at inference time. Note that in the case of Mediapipe framework, we couldn't find the deep learning structure from available resources in the web. From Table~\ref{tab:cmp_param}, it is observable that our proposed framework, compared to the other methods,  has significantly reduced the trainable parameters and instead introduces some time complexity in the inference time, due to using the MDS component. 

From Table~\ref{tab:cmp_bfm}, it is observable that our method has achieved the second rank among the existing methods. It is worth noting here that, compared to the mobileFace, our proposed framework has the advantages of being more efficient in terms of the network trainable parameters (Table~\ref{tab:cmp_param}), and providing an \textit{unbiased} solution, without need to use any 3D model. The other compared methods in this case, like Aldrian's has more complexity in the inference time, along with its biased solution, and Zhao's which only could recover the 3D shape of landmarks from orthographic projection. To validate the results, we performed the wilcoxon signed rank test with significant level $5 \%$ which determines the pairwise $p\-value$ indicating the probability of $x-y$ is from a distribution with zero median. In all cases the pairwise $p-value$ became equal to $8.8 \times 10^{-05}$ which shows all methods' performance is significantly different.

Since there is no expression in the BFM model in this part of the experiments, the resulting 3D shapes may be very similar. Therefore, we performed a trivial case experiment to validate the intelligent behavior of our solutions. To do this, we considered a scenario that our method only outputs one specific solution for all test inputs. In this case the resulting average MSE for BFM dataset became equal to $2.9 \times 10^{-4} $ which is larger than our obtained results in Table~\ref{fig:cmp_bfm}, indicates that our method performs better than this trivial case scenario. Therefore, in the BFM dataset with natural expression, and for pose invariant 3D shape recovery, our framework performs intelligently and comparable with powerful methods with a large number of parameters. We note that the trainable parameters in our deep learning framework are equal to $27505$ which is smaller than the compared methods like MobileFace with about $900000$ parameters and the use of the whole RGB input image in its model-based framework. 
\begin{table}[]
    \centering
    \begin{tabular}{|c|c|c|}
    \hline
         Method & $\#$ Trainable Parameters & Time Complexity (Inference time) \\
         \hline
         Zhao's& $114,840$& $O(1)$\\
         \hline
         Aldrian's & $0$& $O(n^4)$\\
         \hline
         MobileFace & $900,979$ & $O(1) + O($BFM Model$)$\\
         \hline
         RingNet& $> 23000000$& $O(1) + O($FLAME Model$)$\\
         \hline
         Mediapipe &Not found &$O(1)$\\
         \hline
         Ours & $27505$& $O(n^3)$\\
         \hline
    \end{tabular}
    \caption{Number of trainable parameters for the existing methods in our experiments.}
    \label{tab:cmp_param}
\end{table}

\begin{figure}
    \centering
    \includegraphics[scale=0.6]{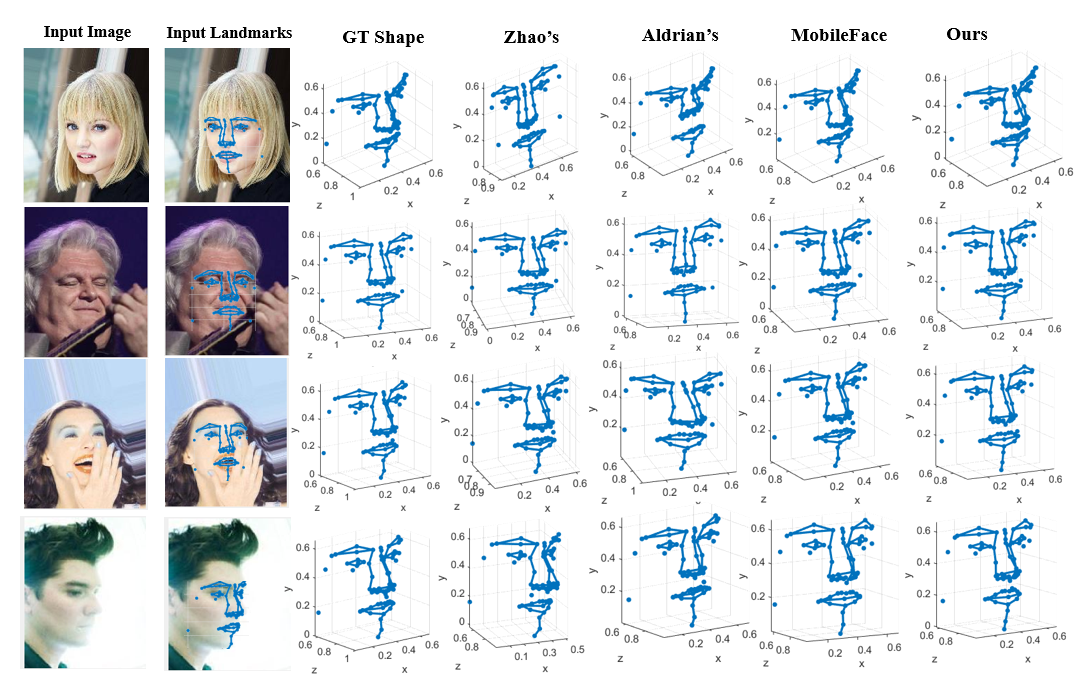}
    \caption{Visual results obtained by existing methods in our experiments on CelebA dataset.}
    \label{fig:cmp_celebA}
\end{figure}

We also used CelebA dataset as real-word human face data for which Figure~\ref{fig:cmp_celebA} illustrates the obtained results by existing methods in our experiments.
In the case of the CelebA dataset with BFM topology, we fitted a 3D model to each image in the CelebA dataset, using the method in~\cite{bas2016fitting} to obtain an estimation of their ground truth (GT) and also the input landmarks. Therefore, we numerically evaluated the methods in this part of the experiments . Table~\ref{tab:cmp_celebA_bfm} include the numerical comparisons using the estimated GT for CelebA dataset, with BFM topology.
\begin{table}
    \centering
    \begin{tabular}{|c|c|c|c|c|}
    \hline
        CelebA-BFM & Zhao's & MobileFace &Aldrian's & Ours \\
        \hline
        avgDepthCorr.($\%$)$\uparrow$ & $71.46 \pm 0.021$ & $71.99 \pm 0.0001$ & $71.91 \pm 0.076$ & $71.99 \pm 0.0084$\\
        \hline
        avgMSE ($\times 10^{-5}$)$\downarrow$ & $21 \pm 0.19$ & $0.092 \pm 0.0016$ & $0.18 \pm 0.012$ & $0.14 \pm 0.037$\\
        \hline
    \end{tabular}
    \caption{Average depth correlation and average MSE obtained by existing methods on CelebA dataset. For each input image BFM landmarks are detected using the method in~\cite{bas2016fitting} and then are used as the test data to evaluate the compared methods in this part of the experiments.}
    \label{tab:cmp_celebA_bfm}
\end{table}
Table~\ref{tab:cmp_celebA_bfm} shows that our framework could perform better than Aldrian's method proposed by the BFM model providers. The different projection types and poses existing in the images make Zhao's method poor in recovering the 3D shape of the input landmarks. Our proposed method is also comparable with the MobileFace framework.
\subsubsection{FLAME Landmark Topology}
FLAME is a 3D model with standard shape and landmark topology on the human faces, having different expressions. In this part of our experiments, we compare our framework with the approaches RingNet~\cite{RingNet:CVPR:2019}, Mediapipe~\cite{lugaresi2019mediapipe}, and Zhao's~\cite{zhao2017simple}, using the FLAME topology.  There are several datasets registered with the FLAME model from which we choose CoMA~\cite{COMA:ECCV18} dataset to train our framework and Zhao's method.

\begin{figure}
    \centering
    \includegraphics[scale=0.7]{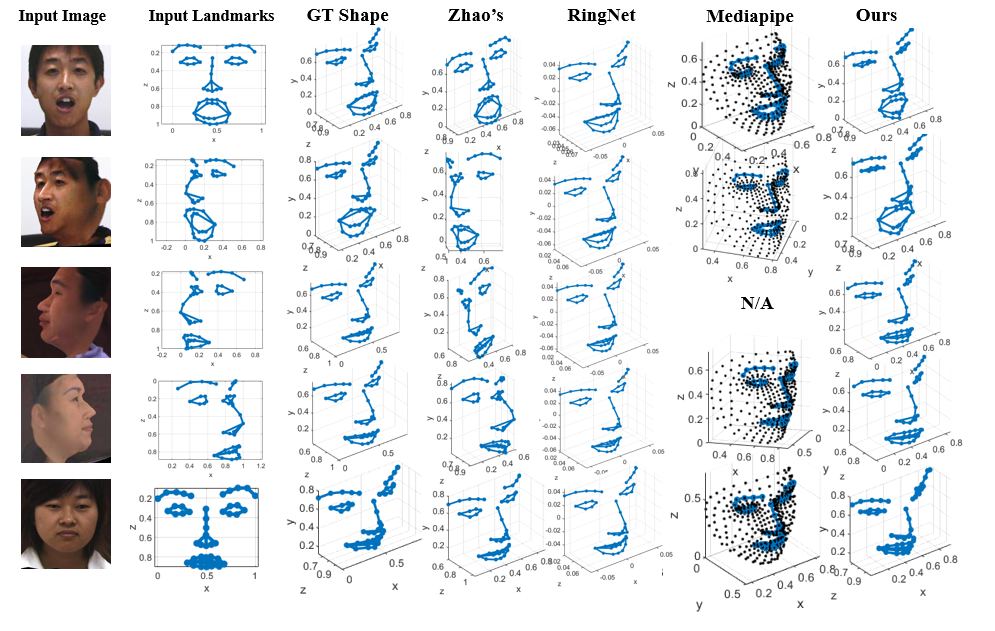}
    \caption{Visual results obtained by our proposed framework and Zhao's work in 3D recovering of the 2D landmarks, based on the FLAME standard for CASIA3D dataset. In the case of the Mediapipe method, the N/A denotes the case for which Mediapipe could not find any solution.}
    \label{fig:cmp_casia3d}
\end{figure}

Figure~\ref{fig:cmp_casia3d} shows the visual result, obtained by the the compared methods in this part of the experiments, for CASIA3D dataset.
From Figure~\ref{fig:cmp_casia3d}, it is observable that our method could suitably recover the 3D expressing shape of the input 2D landmarks, invariant of the pose. Table~\ref{tab:cmp_casia}  includes the numerical comparison for the CASIA3D test dataset to verify the visual results. The Mediapipe method, despite its high accuracy, could not recover the 3D shape of the landmarks in some posed image cases.
\begin{table}[]
    \centering
    \begin{tabular}{|c|c|c|c|c|}
    \hline
        CASIA3D-FLAME & Zhao's&RingNet& Mediapipe & Ours \\
        \hline
        avgDepthCorr.($\%$) $\uparrow$&$60 \pm 0.013$&$65 \pm 0.021$&$68 \pm 0.009$&$65 \pm 0.008$\\
        \hline
         avgMSE ($\times 10^{-3}$)$\downarrow$ & $2.6 \pm 0.30$&$1.34 \pm0.95$& $1.30 \pm 0.19$&$1.50 \pm 0.18$\\
         
         \hline
    \end{tabular}
    \caption{Numerical results obtained by the methods based on FLAME topology on CASIA3D test dataset.}
    \label{tab:cmp_casia}
\end{table}

The numerical results in Table~\ref{tab:cmp_casia} shows that our method achieves comparable performance compared to the Mediapipe and RingNet methods. The wilcoxon p-value for CASIA3D numerical results is equal to $8.8 \times 10^{-05}$ for all pairs of the compared methods indicating their different behavior. 

\begin{figure}
    \centering
    \includegraphics[scale=0.6]{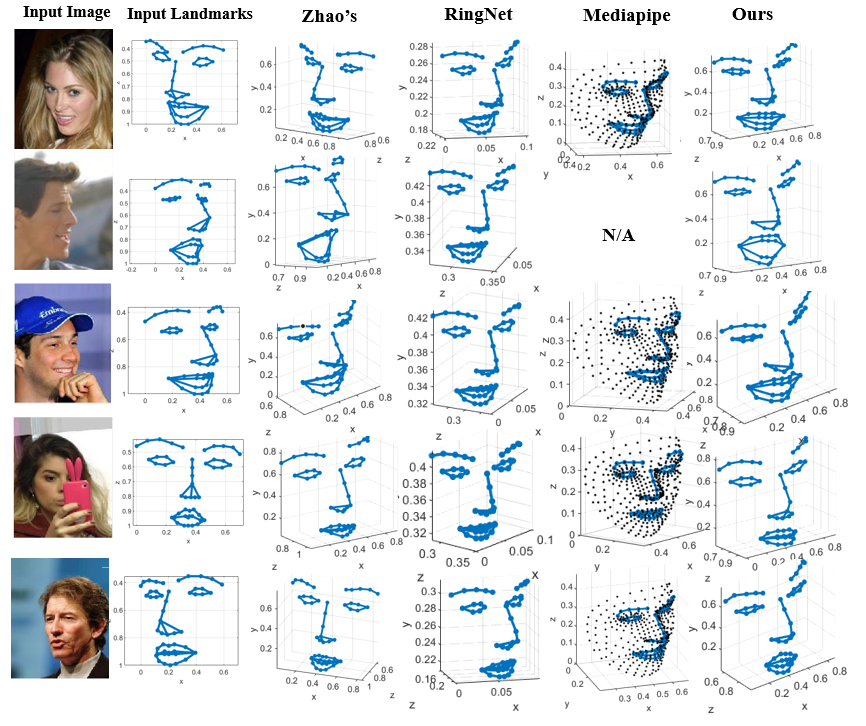}
    \caption{Visual comparison using flame standard landmarks for 3D recovery of landmarks on the faces with expression. }
    \label{fig:cmp_celebA_flame}
\end{figure}

We also report some visual results for CelebA dataset, based on FLAME topology in Figure~\ref{fig:cmp_celebA_flame}, as the real-word evaluation of the existing methods with the FLAME topology in this part of our experiments. 
Figure~\ref{fig:cmp_celebA_flame} also shows suitable 3D shape recovery of 2D landmarks on expressing faces by our method, independent of the projection type.
\subsection{Limitations and Shortcomings}
The idea of our proposed framework is based on using MDS approach as an analytic computing component to reduce the trainable parameters of the proposed framework. In this case, there are some phenomena which affect the performance of the MDS approach in our proposed framework:
\begin{itemize}
    \item Large number of the input landmarks: Since in the MDS approach there is a matrix decomposition step, increasing the number of input landmarks increases the computational burden of our proposed framework.
    \item Intra class diversity among the shapes: this indicates that the MDS approach could not suitably recover the shape of multiple classes of the objects at the same time, and only could be used for sample shapes obeying small dissimilarities. 
\end{itemize}
In general, utilizing the analytical advantages of the MDS approach, e. g., providing an unbiased solution using small number of trainable parameters for the deep learning frameworks, adds some amount of computational complexity in the inference time for the resulting framework.
\subsection{Ablation Study}
The proposed framework in this paper is composed of three components: a deep learning encoder-decoder structure for  turning all input views into a profile view, a deep learning dissimilarity measure to estimate the 3D euclidean distance between the landmarks from input 2D landmarks,  and an MDS component which uses the estimated pairwise 3D euclidean distances for recovering the 3D shape of the input 2D landmarks. It is clear that by removing the encoder-decoder deep learning component, the framework looses its ability of working with different poses and projection types. The MDS component is the central part of the proposed framework which couldn't be modified. Therefore, in the ablation study part of our experiments, we only account for the proposed deep learning dissimilarity, in terms of the selection of the members of a mini batch in its learning phase.

\begin{figure}[h]
    \centering
    \includegraphics[scale=0.5]{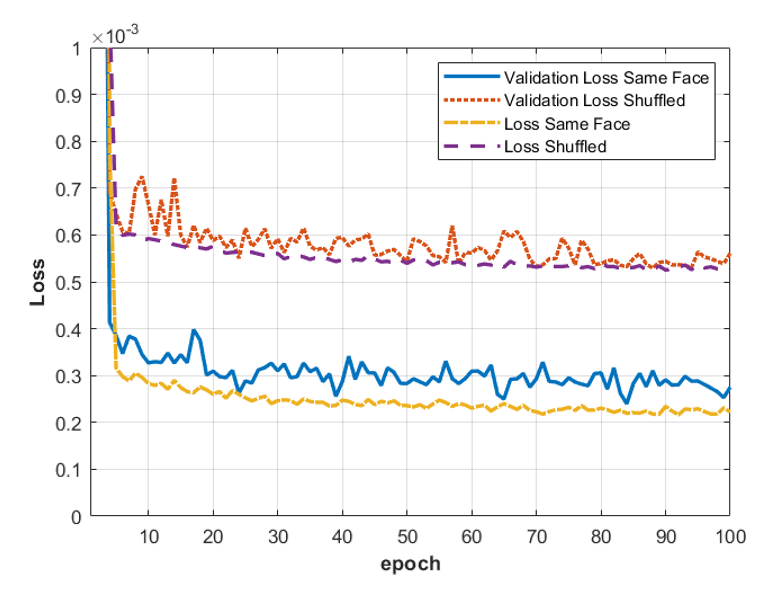}
    \caption{Comparison of two training schemes for the proposed dissimilarity measure, for 
    BFM dataset. The used schemes include using the pairs of input 2D landmarks from the same human face as the mini-batch members and shuffled pairs from different human faces.}
    \label{fig:ablation}
\end{figure}

Since our proposed deep learning dissimilarity only accepts two single points as the input, the mini-batch members in the training process could be selected from the \underline{pairs} of  2D points in the \textit{same} human face, or from different faces. To see whether the selection of the pairs of 2D landmarks in a mini-batch affect the learning process of the proposed dissimilarity or not, we compared two training schemes with mini-batches including the landmark pairs from the same person, and mini-batches including shuffled landmark pairs from different human faces, for BFM dataset. Figure~\ref{fig:ablation} shows the learning curve for the above training schemes, from which we observe that the selection of the landmark pairs from the same face in the training process improves the accuracy of the obtained results.

\section{Conclusions and Future Works}
\label{sec:conclusion}
In this paper, we propose a low parameter deep learning framework to recover the 3D locations of 2D landmarks on the human face in a single input image, independent of the type of image formation like pose, projection type, etc. In our proposed framework, we utilize the MDS approach for mapping the 2D points into 3D shape space preserving their configuration, by learning the 3D Euclidean distance from 2D landmark distances resulting in a symmetric dissimilarity in the non-metric MDS approach. 

Compared to similar methods from the literature, the performance evaluation of our proposed framework verifies its plausibility in pose and projection type invariant 3D landmark shape recovery, which results in plausible face shapes. The proposed framework could be incorporated into a high-resolution 3D human face reconstruction framework or in mobile devices, as a low parameter component with promising accuracy to increase the quality of 3D reconstructed faces. 

The future works may concentrate on using the proposed method for objects other than the human face by resolving the self-occlusion phenomenon about the target object of interest, and investigating the solutions for the existing limitations of  proposed framework like designing a more complex dissimilarity structure capable of recovering the landmarks' 3D shape from different poses and projection types, by itself, and accordingly removing the need to use the encoder-decoder component in the framework.
\section{Acknowledgements}
We are also immensely grateful to Professor Ali Ghodsi for their comments on an earlier version of the manuscript, although any errors are our own and should not tarnish the reputations of these esteemed persons.

\bibliographystyle{unsrtnat}
\bibliography{template.bib}  






\end{document}